\documentclass[11pt]{article}
\usepackage{authblk}
\usepackage{eacl2017}
\usepackage[utf8]{inputenc} 
\usepackage[T1]{fontenc}    
\usepackage{hyperref}       
\usepackage{url}            
\usepackage{booktabs}       
\usepackage{amsfonts}       
\usepackage{nicefrac}       
\usepackage{microtype}      
\usepackage{times}
\usepackage{latexsym}
\usepackage{epsfig}
\usepackage{array}
\usepackage{color}
\usepackage{multirow}
\usepackage{subfig}
\usepackage{algorithmic}
\usepackage{algorithm}
\usepackage{multirow}
\usepackage{rotating}
\usepackage{booktabs}
\usepackage{makecell}
\usepackage{comment}
\usepackage{amsmath,amsfonts,amssymb}
\usepackage{tikz}
\usepackage{verbatim}
\usepackage{caption}
\usepackage{booktabs}
\usepackage{ctable}
\usepackage{makecell}
\usepackage{array} 
\usepackage{varwidth} 
\usepackage{stmaryrd}

\newcommand{\argmax}{\operatornamewithlimits{argmax}}
\newcommand*{\Scale}[2][4]{\scalebox{#1}{$#2$}}%

\eaclfinalcopy 
\def\eaclpaperid{3} 

\title{A Network-based End-to-End Trainable Task-oriented Dialogue System}

\author[1]{\bf Tsung-Hsien Wen}
\author[1]{\bf David Vandyke}
\author[1]{\bf Nikola Mrk{\v{s}}i\'c}
\author[1]{\bf Milica Ga{\v{s}}i\'c}
\author[1]{\\ \bf Lina M. Rojas-Barahona}
\author[1]{\bf Pei-Hao Su}
\author[1]{\bf Stefan Ultes}
\author[1]{\bf Steve Young}
\affil[1]{Cambridge University Engineering Department,}
\affil[ ]{Trumpington Street, Cambridge, CB2 1PZ, UK}
\affil[ ]{\tt \{thw28,djv27,nm480,mg436,lmr46,phs26,su259,sjy11\}@cam.ac.uk}

\begin{document}
\maketitle
\begin{abstract}
Teaching machines to accomplish tasks by conversing naturally with humans is challenging. Currently, developing task-oriented dialogue systems requires creating multiple components and typically this involves either a large amount of handcrafting, or acquiring costly labelled datasets to solve a statistical learning problem for each component. In this work we introduce a neural network-based text-in, text-out end-to-end trainable goal-oriented dialogue system along with a new way of collecting dialogue data based on a novel pipe-lined Wizard-of-Oz framework. This approach allows us to develop dialogue systems easily and without making too many assumptions about the task at hand. The results show that the model can converse with human subjects naturally whilst helping them to accomplish tasks in a restaurant search domain.
\end{abstract}

\section{Introduction}\label{sec:intro}

Building a task-oriented dialogue system such as a hotel booking or a technical support service is difficult because it is application-specific and there is usually limited availability of training data.
To mitigate this problem, recent machine learning approaches to task-oriented dialogue system design have cast the problem as a partially observable Markov Decision Process (POMDP)~\cite{6407655} with the aim of using reinforcement learning (RL) to train dialogue policies online through interactions with real users~\cite{6639297}.
However, the language understanding~\cite{henderson14,export228844} and language generation~\cite{wensclstm15,wenmultinlg16} modules still rely on supervised learning and therefore need corpora to train on.
Furthermore, to make RL tractable, the state and action space must be carefully designed~\cite{6407655,Young2010HIS}, which may restrict the expressive power and  learnability of the model.
Also, the reward functions needed to train such models are difficult to design and hard to measure at run-time~\cite{SuVGKMWY15,su2016acl}. 

At the other end of the spectrum, sequence to sequence learning~\cite{SutskeverVL14} has inspired several efforts to build end-to-end trainable, non-task-oriented conversational systems~\cite{VinyalsL15,ShangLL15,SerbanSBCP15}.
This family of approaches treats dialogue as a source to target sequence transduction problem, applying an encoder network~\cite{ChoMGBSB14} to encode a user query into a distributed vector representing its semantics, which then conditions a decoder network  to generate each system response.
These models typically require a large amount of data to train.  They allow the creation of effective chatbot type systems but they lack any capability for supporting domain specific tasks, for example, being able to interact with databases~\cite{Sukhbaatar2015EndToEndMN,YinLLK15} and aggregate useful information into their responses.

\begin{figure*}[t]
\centerline{\includegraphics[width=100mm]{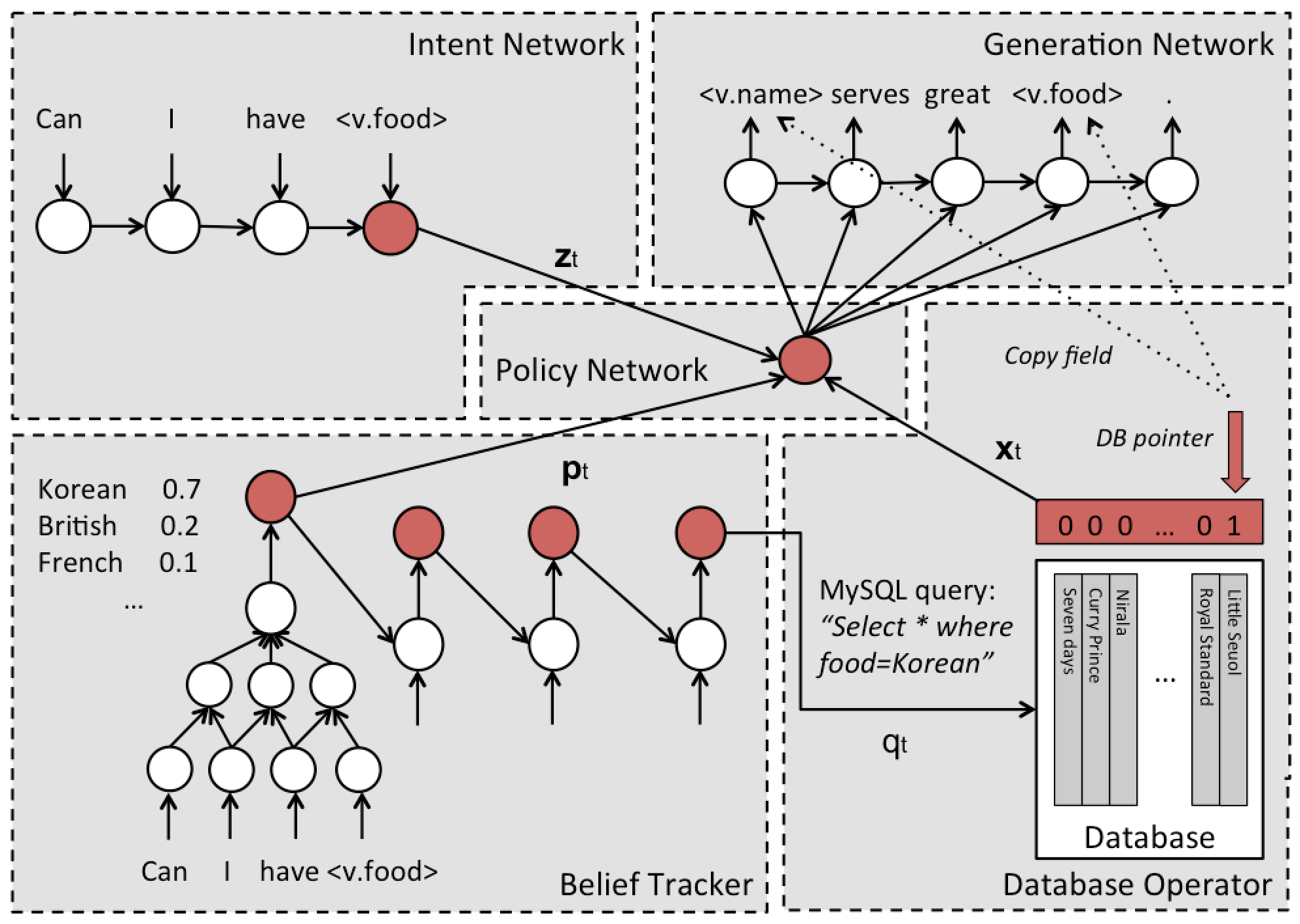}}
\caption{The proposed end-to-end trainable dialogue system framework}
\label{fig:n2n}
\vspace{-3mm}
\end{figure*}

In this work, we propose a neural network-based model for task-oriented dialogue systems by balancing the strengths and the weaknesses of the two research communities:
the model is end-to-end trainable\footnote{We define {\it end-to-end trainable} as that each system module is trainable from data except for a database operator.} but still modularly connected;
it does not directly model the user goal, but nevertheless, it still learns to accomplish the required task by providing  {\it relevant} and {\it appropriate} responses at each turn;
it has an explicit representation of database (DB) attributes (slot-value pairs) which it uses to achieve a high task success rate, but has a distributed representation of user intent (dialogue act) to allow ambiguous inputs; and
it uses delexicalisation\footnote{\label{fn:delex}Delexicalisation: we replaced slots and values by generic tokens (e.g. keywords like Chinese or Indian are replaced by <v.food> in Figure~\ref{fig:n2n}) to allow weight sharing.} and a weight tying strategy~\cite{henderson14} to reduce the data required to train the model, but still maintains a high degree of freedom should larger amounts of data become available.
We show that the proposed model performs a given task very competitively across several metrics when trained on only a few hundred dialogues.

In order to train the model for the target application, we introduce a novel pipe-lined data collection mechanism inspired by the Wizard-of-Oz paradigm~\cite{Kelley84} to collect human-human dialogue corpora via crowd-sourcing.
We found that this process is simple and enables fast data collection online with very low development costs.

\section{Model}\label{sec:model}

We treat dialogue as a sequence to sequence mapping problem (modelled by a sequence-to-sequence architecture~\cite{SutskeverVL14}) augmented with the dialogue history (modelled by a set of belief trackers~\cite{henderson14}) and the current database search outcome (modelled by a database operator), as shown in Figure~\ref{fig:n2n}.
At each turn, the system takes a sequence of tokens\textsuperscript{\ref{fn:delex}} from the user as input and converts it into two internal representations: 
a distributed representation generated by an intent network and a probability distribution
over slot-value pairs called the belief state~\cite{6407655} generated by a set of belief trackers.
The database operator then selects the most probable values in the belief state to form a query to the DB, and  
the search result, along with the intent representation and belief state are transformed and combined by a policy network to form a single vector representing the next system action.
This system action vector is then used to condition a response generation network~\cite{wenrgm15,wensclstm15} which generates the required system output token by token in skeletal form.
The final system response is then formed by substituting the actual values of the database entries into the skeletal sentence structure. A more detailed description of each component is given below.

\begin{figure*}[t]
\centerline{\includegraphics[width=120mm]{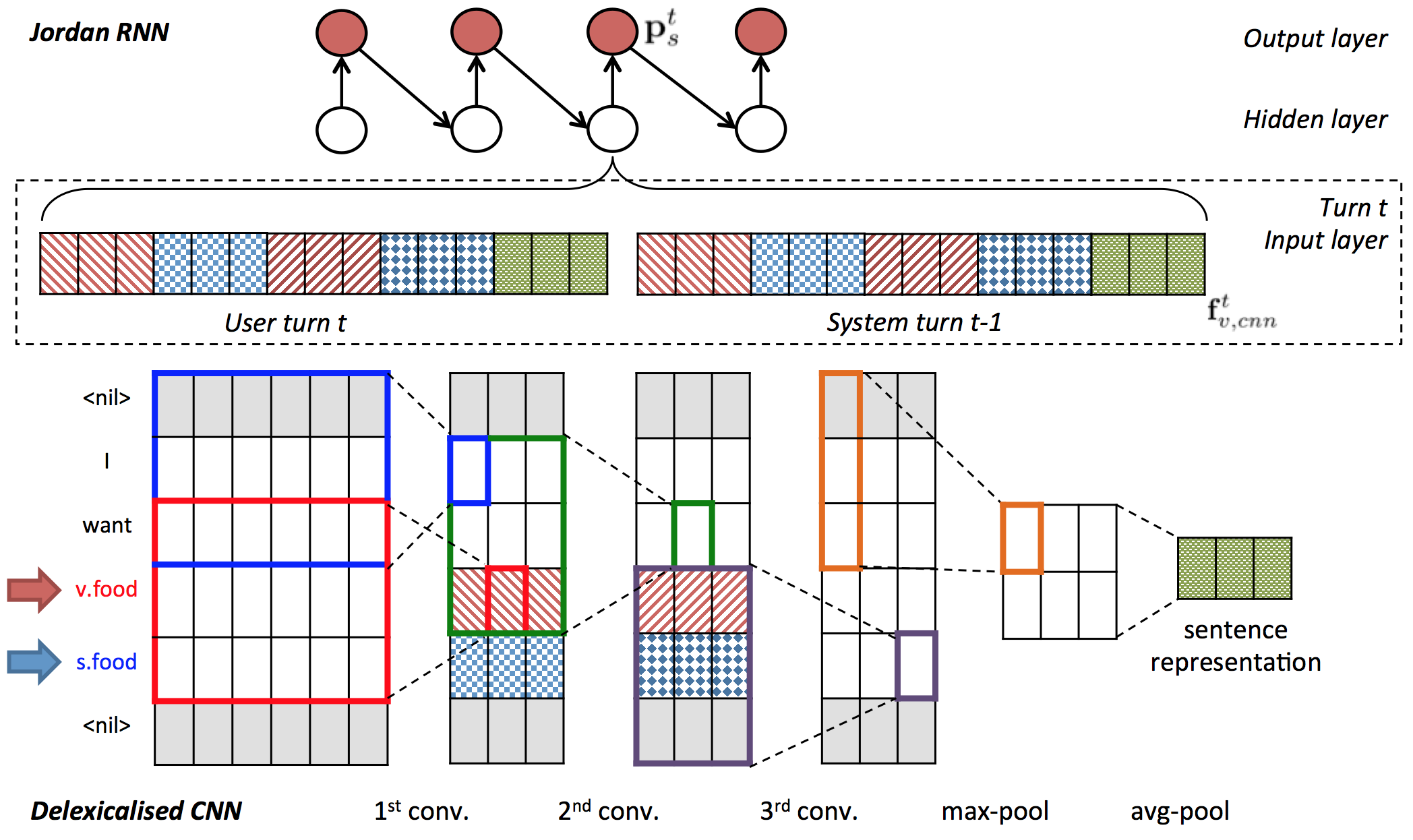}}
\caption{Tied Jordan-type RNN belief tracker with delexicalised CNN feature extractor. The output of the CNN feature extractor is a concatenation of top-level sentence (green) embedding and several levels of intermediate ngram-like embeddings (red and blue). However, if a value cannot be delexicalised in the input, its ngram-like embeddings will all be padded with zeros. We pad zero vectors (in gray) before each convolution operation to make sure the representation at each layer has the same length. The output of each tracker $\mathrm{\mathbf{p}}_{s}^{t}$ is a distribution over values of a particular slot $s$.}
\label{fig:tracker}
\vspace{-3mm}
\end{figure*}

\subsection{Intent Network}~\label{ssec:encoder}

The intent network can be viewed as the encoder in the sequence-to-sequence learning framework~\cite{SutskeverVL14} whose job is to encode a sequence of input tokens $w_{0}^t,w_{1}^t,...w_{N}^t$ into a distributed vector representation $\mathrm{\mathbf{z}}_{t}$ at every turn $t$.
Typically, a Long Short-term Memory (LSTM) network~\cite{Hochreiter1997} is used and the last time step hidden layer $\mathrm{\mathbf{z}}_t^{N}$ is taken as the representation,
\begin{equation}
\mathrm{\mathbf{z}}_t = \mathrm{\mathbf{z}}_{t}^{N} = \text{LSTM}(w_{0}^t,w_{1}^t,...w_{N}^t)
\end{equation}
Alternatively, a convolutional neural network (CNN) can be used in place of the LSTM as the encoder ~\cite{KalchbrennerGB14,kim2014},
\begin{equation}
\mathrm{\mathbf{z}}_t = \text{CNN}(w_{0}^t,w_{1}^t,...w_{N}^t)
\end{equation}
and here we investigate both.
Since all the slot-value specific information is delexicalised, the encoded vector can be viewed as a distributed intent representation which replaces the hand-coded dialogue act representation~\cite{Traum1999} in traditional task-oriented dialogue systems. 

\subsection{Belief Trackers}~\label{ssec:tracker}
\vspace{-3mm}

Belief tracking (also called Dialogue State tracking) provides the core of a task-oriented spoken dialogue system (SDS)~\cite{Henderson2015a}.
Current state-of-the-art belief trackers use discriminative models such as recurrent neural networks (RNN)~\cite{39298195,WenHLTL13} to directly map ASR hypotheses to belief states~\cite{henderson14,mrksic2016nbt}.
Although in this work we focus on text-based dialogue systems, we retain belief tracking at the core of our system because: (1) it enables a sequence of free-form natural language sentences to be mapped into a fixed set of slot-value pairs, which can then be used to query a DB. This can be viewed as a simple version of a semantic parser~\cite{Berant2013SemanticPO}; (2) by keeping track of the dialogue state, it avoids learning unnecessarily complicated long-term dependencies from raw inputs; (3) it uses a smart weight tying strategy that can greatly reduce the data required to train the model, and (4) it provides an inherent robustness which simplifies future extension to spoken systems.

Using each user input as new evidence, the task of a belief tracker is to maintain a multinomial distribution $p$ over values $v \in V_{s}$ for each informable slot $s$, and a binary distribution for each requestable slot\footnote{\label{fn:slots}Informable slots are slots that users can use to constrain the search, such as food type or price range; Requestable slots are slots that users can ask a value for, such as address.}.
Each slot in the ontology $\mathbb{G}$\footnote{A small knowledge graph defining the slot-value pairs the system can talk about for a particular task.} has its own specialised tracker, and each tracker is a Jordan-type (recurrence from output to hidden layer)~\cite{jordan89} RNN\footnote{We don't use the recurrent connection for requestable slots since they don't need to be tracked.} with a CNN feature extractor, as shown in Figure~\ref{fig:tracker}.
Like~\newcite{Mrksic15}, we tie the RNN weights together for each value $v$ but vary features $\mathrm{\mathbf{f}}_{v}^{t}$ when updating each pre-softmax activation $g_{v}^{t}$. The update equations for a given slot $s$ are,
\begin{alignat}{3}
\mathrm{\mathbf{f}}_{v}^{t} &= \mathrm{\mathbf{f}}_{v, cnn}^{t} \oplus p_{v}^{t-1} \oplus p_{\emptyset}^{t-1}\\
g_{v}^{t} &= \mathrm{\mathbf{w}}_s \cdot \text{sigmoid}(\mathrm{\mathbf{W}}_{s}\mathrm{\mathbf{f}}_v^{t}+\mathrm{\mathbf{b}}_{s}) + b'_{s}\\
p_{v}^{t} &= \frac{\text{exp}(g_{v}^{t})}{\text{exp}(g_{\emptyset,s})+\sum_{v'\in V_{s}}\text{exp}(g_{v'}^{t})}\label{eq:trkp}
\end{alignat}
where vector $\mathrm{\mathbf{w}}_s$, matrix $\mathrm{\mathbf{W}}_{s}$, bias terms $\mathrm{\mathbf{b}}_{s}$ and $b'_{s}$, and scalar $g_{\emptyset,s}$ are parameters. 
$p_{\emptyset}^{t}$ is the probability that the user has not mentioned that slot up to turn $t$ and can be calculated by substituting $g_{\emptyset,s}$ for $g_{v}^t$ in the numerator of Equation~\ref{eq:trkp}.			
In order to model the discourse context at each turn, the feature vector $\mathrm{\mathbf{f}}_{v, cnn}^{t}$ is the concatenation of two CNN derived features, one from processing the user input $u_{t}$ at turn $t$ and the other from processing the machine response $m_{t-1}$ at turn $t-1$, 
\begin{equation}
\mathrm{\mathbf{f}}_{v, cnn}^{t} = \text{CNN}^{(u)}_{s,v}(u_t) \oplus \text{CNN}^{(m)}_{s,v}(m_{t-1})
\end{equation}
where every token in $u_t$ and $m_{t-1}$ is represented by an embedding of size $N$ derived from a 1-hot input vector.
In order to make the tracker aware when delexicalisation is applied to a slot or value, the slot-value specialised CNN operator $\text{CNN}^{(\cdot)}_{s,v}(\cdot)$ extracts not only the top level sentence representation but also intermediate n-gram-like embeddings determined by the position of the delexicalised token in each utterance. 
If multiple matches are observed, the corresponding embeddings are summed.
On the other hand, if there is no match for a particular slot or value, the empty n-gram embeddings are padded with zeros.
In order to keep track of the position of delexicalised tokens, both sides of the sentence are padded with zeros before each convolution operation. 
The number of vectors is determined by the filter size at each layer.
The overall process of extracting several layers of position-specific features is visualised in Figure~\ref{fig:tracker}.

The belief tracker described above is based on~\newcite{henderson14} with some modifications: (1) only probabilities over informable and requestable slots and values are output, (2) the recurrent memory block is removed, since it appears to offer no benefit in this task, and (3) the n-gram feature extractor is replaced by the CNN extractor described above.
By introducing slot-based belief trackers, we essentially add a  set of intermediate labels into the system as compared to training a pure end-to-end system.
Later in the paper we will show that these tracker components are critical for achieving task success.  We will also show that the additional annotation requirement that they introduce can be  successfully mitigated using a novel pipe-lined Wizard-of-Oz data collection framework.

\subsection{Policy Network and Database Operator}~\label{spec:policy}
\noindent
{\bf Database Operator} \hspace{2mm} Based on the output $\mathrm{\mathbf{p}}_{s}^{t}$ of the belief trackers, the DB query $q_{t}$ is formed by,
\begin{equation}
q_{t} = \bigcup_{s'\in S_{I}}\{\argmax_{v} \mathrm{\mathbf{p}}_{s'}^{t}\}
\end{equation}
where $S_{I}$ is the set of informable slots.
This query is then applied to the DB to create a binary truth value vector
$\mathrm{\mathbf{x}}_{t}$ over DB entities where a 1 indicates that the corresponding 
entity is consistent with the query (and hence it is consistent with the most likely belief state). 
In addition, if $\mathbf{x}$ is not entirely null, an associated entity pointer is maintained which identifies one of the matching entities selected at random.  
The entity pointer is updated if the current entity no longer matches the search criteria; otherwise it stays the same.
The entity referenced by the entity pointer is used to form the final system response as described in Section \ref{ssec:gen}.

{\bf Policy network} \hspace{2mm} The policy network can be viewed as the glue which binds the system modules together. Its output is a single vector $\mathrm{\mathbf{o}}_t$ representing the system action, and its inputs are comprised of $\mathrm{\mathbf{z}}_t$ from the intent network, the belief state $\mathrm{\mathbf{p}}_{s}^{t}$, and the DB truth value vector $\mathrm{\mathbf{x}}_t$.
Since the generation network only generates appropriate sentence forms, the individual probabilities of the categorical values in the informable belief state are immaterial and are summed together to form a summary belief vector for each slot $\mathrm{\mathbf{\hat{p}}}_{s}^{t}$ represented by three components: the summed value probabilities, the probability that the user said they "don't care" about this slot and the probability that the slot has not been mentioned. 
Similarly for the truth value vector $\mathrm{\mathbf{x}}_t$, the number of matching entities matters but not their identity. This vector is therefore compressed to a 6-bin 1-hot encoding $\mathrm{\mathbf{\hat{x}}}_{t}$, which represents different degrees of matching in the DB (no match, 1 match, ... or more than 5 matches).
Finally, the policy network output is generated by a three-way matrix transformation,
\begin{equation}
\mathrm{\mathbf{o}}_t = \tanh(\mathrm{\mathbf{W}}_{zo}\mathrm{\mathbf{z}}_{t} + \mathrm{\mathbf{W}}_{po}\mathrm{\mathbf{\hat{p}}}_{t} + \mathrm{\mathbf{W}}_{xo}\mathrm{\mathbf{\hat{x}}}_{t})
\end{equation}
where matrices $\mathrm{\mathbf{W}}_{zo}$, $\mathrm{\mathbf{W}}_{po}$, and $\mathrm{\mathbf{W}}_{xo}$ are parameters and $\mathrm{\mathbf{\hat{p}}}_{t} = \bigoplus_{s\in \mathbb{G}}\mathrm{\mathbf{\hat{p}}}^{t}_{s}$ is a concatenation of all summary belief vectors.

\subsection{Generation Network}~\label{ssec:gen}\vspace{-3mm}

The generation network uses the action vector $\mathrm{\mathbf{o}}_t$ to condition a language generator~\cite{wensclstm15}.  This generates template-like sentences token by token based on the language model probabilities,
\begin{equation}
P(w_{j+1}^{t}|w_{j}^{t},\mathrm{\mathbf{h}}_{j-1}^{t},\mathrm{\mathbf{o}}_{t}) = \text{LSTM}_{j}(w_{j}^{t},\mathrm{\mathbf{h}}_{j-1}^{t},\mathrm{\mathbf{o}}_{t})
\end{equation}
where $\text{LSTM}_{j}(\cdot)$ is a conditional LSTM operator for one output step $j$, $w_{j}^{t}$ is the last output token (i.e. a word, a delexicalised slot name or a delexicalised slot value), and $\mathrm{\mathbf{h}}_{j-1}^{t}$ is the hidden layer.
Once the output token sequence has been generated, the generic tokens are replaced by their actual values: (1) replacing delexicalised slots by random sampling from a list of surface forms, e.g. {\it <s.food>} to {\it food} or {\it type of food}, and (2) replacing delexicalised values by the actual attribute values of the entity currently selected by the DB pointer.
This is similar in spirit to the Latent Predictor Network~\cite{Wang16lpn} where the token generation process is augmented by a set of pointer networks~\cite{NIPS2015_5866} to transfer entity specific information into the response.

{\bf Attentive Generation Network} \hspace{2mm} Instead of decoding responses directly from a static action vector $\mathrm{\mathbf{o}}_t$, an attention-based mechanism~\cite{BahdanauCB14,HermannKGEKSB15} can be used to dynamically aggregate source embeddings at each output step $j$.
In this work we explore the use of an attention mechanism to combine the tracker belief states, i.e.\  $\mathrm{\mathbf{o}}_t$ is computed at each output step $j$ by,
\begin{equation}
\mathrm{\mathbf{o}}_t^{(j)} = \tanh(\mathrm{\mathbf{W}}_{zo}\mathrm{\mathbf{z}}_{t} + \mathrm{\mathbf{\hat{p}}}_{t}^{(j)} + \mathrm{\mathbf{W}}_{xo}\mathrm{\mathbf{\hat{x}}}_{t})
\end{equation}
where for a given ontology $\mathbb{G}$,
\begin{equation}
\mathrm{\mathbf{\hat{p}}}_{t}^{(j)} = \sum_{s\in \mathbb{G}} \alpha_{s}^{(j)} \tanh(\mathrm{\mathbf{W}}_{po}^{s} \cdot \mathrm{\mathbf{\hat{p}}}_{s}^{t})
\end{equation}
and where the attention weights $\alpha_{s}^{(j)}$ are calculated by a scoring function, 
\begin{equation}
\alpha_{s}^{(j)} = \text{softmax}\big( \mathrm{\mathbf{r}}^\intercal \tanh( \mathrm{\mathbf{W}}_{r} \cdot \mathrm{\mathbf{u}}_{t} ) \big)
\end{equation}
where $\mathrm{\mathbf{u}}_{t}=\mathrm{\mathbf{z}}_{t} \oplus \mathrm{\mathbf{\hat{x}}}_{t} \oplus \mathrm{\mathbf{\hat{p}}}_{s}^{t} \oplus \mathrm{\mathbf{w}}_{j}^{t} \oplus \mathrm{\mathbf{h}}_{j-1}^{t}, $ matrix $\mathrm{\mathbf{W}}_{r}$, and vector $\mathrm{\mathbf{r}}$ are parameters to learn and $\mathrm{\mathbf{w}}_{j}^{t}$ is the embedding of token $w_{j}^{t}$.

\begin{table*}[t]
  \caption{Tracker performance in terms of Precision, Recall, and F-1 score.}
  \vspace{-2mm}
  \label{tab:tracker}
  \centering
  \begin{tabular}{lcccccc}
    \toprule
    \multirow{2}{1.5cm}{Tracker type}	&\multicolumn{3}{c}{Informable}&\multicolumn{3}{c}{Requestable}\\
    \cmidrule{2-7}
   		&	Prec. 	& 	Recall     	& 	F-1		&	Prec.     	& 	Recall     	&	F-1 		\\
    \midrule
    cnn	&	99.77\%	&	96.09\%	&	97.89\%	&	98.66\%	&	93.79\%	&	96.16\%	\\
    ngram	&	99.34\%	&	94.42\%	&	96.82\%	&	98.56\%	&	90.14\%	&	94.16\%	\\
    \bottomrule
  \end{tabular}
  \vspace{-3mm}
\end{table*}

\section{Wizard-of-Oz Data Collection}\label{sec:woz}

Arguably the greatest bottleneck for statistical approaches to dialogue system development is the collection of appropriate training data, and this is especially true for task-oriented dialogue systems.
Serban et al~\cite{SerbanLCP15} have catalogued existing corpora for developing conversational agents.  Such corpora may be useful for bootstrapping, but, for
task-oriented dialogue systems, in-domain data is essential\footnote{E.g. technical support for Apple computers may  differ completely from that for Windows, due to the many differences in software and hardware.}. 
To mitigate this problem,  we propose a novel crowdsourcing version of the Wizard-of-Oz (WOZ) paradigm~\cite{Kelley84} for collecting domain-specific corpora. 

Based on the given ontology, we designed two webpages on Amazon Mechanical Turk, one for wizards and the other for users (see Figure~\ref{fig:wozu} and~\ref{fig:wozw} for the designs).
The users are given a task specifying the characteristics of a particular entity that they must find (e.g. {\it a Chinese restaurant in the north}) and asked to type in natural language sentences to fulfil the task. 
The wizards are given a form to record the information conveyed in the last user turn (e.g. {\it pricerange=Chinese, area=north}) and a search table showing all the available matching entities in the database.
Note these forms contain all the labels needed to train the slot-based belief trackers.
The table is automatically updated every time the wizard submits new information.
Based on the updated table, the wizard types an appropriate system response and the dialogue continues.

In order to enable large-scale parallel data collection and avoid the distracting latencies inherent in conventional WOZ scenarios~\cite{bohus08}, users and wizards are asked to contribute just a single turn to each dialogue.
To ensure coherence and consistency, users and wizards must review all previous turns in that dialogue before they contribute their turns.  Thus dialogues progress in a pipe-line.  Many dialogues can be active in parallel and no worker ever has to wait for a response from the other party in the dialogue.
Despite the fact that multiple workers contribute to each dialogue, we observe that dialogues are generally coherent yet diverse.  Furthermore, this 
turn-level data collection strategy seems to encourage workers to learn and correct each other based on previous turns.

In this paper, the system was designed to assist users to find a restaurant in the Cambridge, UK area. 
There are three informable slots ({\it food}, {\it pricerange}, {\it area}) that users can use to constrain the search and six requestable slots ({\it address}, {\it phone}, {\it postcode} plus the three informable slots) that the user can ask a value for once a restaurant has been offered. There are 99 restaurants in the DB.
Based on this domain, we ran 3000 HITs (Human Intelligence Tasks) in total for roughly 3 days and collected 1500 dialogue turns.
After cleaning the data, we have approximately 680 dialogues in total (some of them are unfinished).
The total cost for collecting the dataset was $\sim400$ USD.

\section{Empirical Experiments}\label{sec:corpus_eval}

{\bf Training} \hspace{2mm}
Training is divided into two phases. 
Firstly the belief tracker parameters $\theta_{b}$ are trained using the cross entropy errors between tracker labels $\mathrm{\mathbf{y}}_{s}^{t}$ and predictions $\mathrm{\mathbf{p}}_{s}^{t}$,
$L_{1}(\theta_{b}) = -\sum_{t}\sum_{s} (\mathrm{\mathbf{y}}_{s}^{t})^\intercal \log \mathrm{\mathbf{p}}_{s}^{t}$.
For the full model, we have three informable trackers ({\it food, pricerange, area}) and seven requestable trackers ({\it address, phone, postcode, name}, plus the three informable slots).

Having fixed the tracker parameters, the remaining parts of the model $\theta_{\backslash b}$ are trained using the cross entropy errors from the generation network language model,
$L_{2}(\theta_{\backslash b}) = -\sum_{t}\sum_{j} (\mathrm{\mathbf{y}}_{j}^{t})^\intercal \log \mathrm{\mathbf{p}}_{j}^{t}$,
where $\mathrm{\mathbf{y}}_{j}^{t}$ and $\mathrm{\mathbf{p}}_{j}^{t}$ are output token targets and predictions respectively, at turn $t$ of output step $j$.
We treated each dialogue as a batch and used stochastic gradient decent with a small $l2$ regularisation term to train the model.
The collected corpus was partitioned into a training, validation, and testing sets in the ratio 3:1:1. 
Early stopping was implemented based on the validation set for regularisation and
gradient clipping was set to 1.
All the hidden layer sizes were set to 50, and all the weights were randomly initialised between -0.3 and 0.3 including word embeddings.
The vocabulary size is around 500 for both input and output, in which rare words and words that can be delexicalised are removed.
We used three convolutional layers for all the CNNs in the work and all the filter sizes were set to 3. 
Pooling operations were only applied after the final convolution layer.

{\bf Decoding} \hspace{2mm} In order to decode without length bias, we decoded each system response $m_t$ based on the average log probability of tokens,
\begin{equation}\label{eq:ml}
 m_t^* = \argmax_{m_t}\{ \log p(m_t|\theta, u_t)/J_{t} \}
\end{equation}
where $\theta$ are the model parameters, $u_t$ is the user input, and $J_{t}$ is the length of the machine response.

As a contrast, we also investigated the MMI criterion~\cite{LiGBGD15} to increase diversity and put additional scores on delexicalised tokens to encourage task completion.
This {\it weighted} decoding strategy has the following objective function,
\begin{align}\label{eq:weighted}
 m_t^* = 	\argmax_{m_t}\{&\log p(m_t|\theta,u_t)/J_t - \\ \nonumber
 					&\lambda \log p(m_t)/J_t + \gamma R_t  \}
\end{align}
where $\lambda$ and $\gamma$ are weights selected on validation set and $ \log p(m_t)$ can be modelled by a standalone LSTM language model.
We used a simple heuristic for the scoring function $R_t$ designed to reward giving appropriate information and penalise spuriously providing unsolicited information\footnote{We give an additional reward if a requestable slot (e.g. address) is requested and its corresponding delexicalised slot or value token (e.g. <v.address> and <s.address>) is generated. We give an additional penalty if an informable slot is never mentioned (e.g. food=none) but its corresponding delexicalised value token is generated (e.g. <v.food>). For more details on scoring, please see Table~\ref{tab:R}.}.
We applied beam search with a beamwidth equal to 10, the search stops when an end of sentence token is generated.
In order to obtain language variability from the deployed model we ran decoding until we obtained 5 candidates and randomly sampled one as the system response.

{\bf Tracker performance} \hspace{2mm} Table~\ref{tab:tracker} shows the evaluation of the trackers' performance. 
Due to delexicalisation, both CNN type trackers and N-gram type trackers~\cite{henderson14} achieve high precision, but the N-gram tracker has worse recall. 
This result suggests that compared to simple N-grams, CNN type trackers can better generalise to sentences with long distance dependencies and more complex syntactic structures.

\begin{table*}[t]
  \caption{Performance comparison of different model architectures based on a corpus-based evaluation.}
  \vspace{-2mm}
  \label{tab:main}
  \centering
  \Scale[0.85]{
  \begin{tabular}{llllcccc}
    \toprule
    &		Encoder	&	Tracker	&	Decoder	&	Match(\%)	&	Success(\%)	&	T5-BLEU	&	T1-BLEU\\
    \midrule
    \multicolumn{7}{l}{\bf Baseline}\\
    &		lstm		&	-	 	& 	lstm	     	& 	-		&	-     			& 	0.1650   	&	0.1718 \\
    &		lstm		&	turn recurrence  & 	lstm	& 	-		&	-     			& 	0.1813     	&	0.1861 \\
    \midrule
    \multicolumn{7}{l}{\bf Variant}\\
    &		lstm		&	rnn-cnn, w/o req.&	lstm	&	89.70	&	30.60		&	0.1769 	&	0.1799 \\
    &   	cnn		&	rnn-cnn	&	lstm		&	88.82	&	58.52		&	0.2354 	&	0.2429 \\
    \midrule
    \multicolumn{7}{l}{\bf Full model w/ different decoding strategy}\\
    &    	lstm		&	rnn-cnn	&	lstm				&	86.34	&	75.16		&	0.2184 	&	0.2313 \\
    &		lstm		&	rnn-cnn	&	+ weighted 		&	86.04	&	78.40		&	0.2222	&	0.2280 \\
    &		lstm		&	rnn-cnn	&	+ att.				&	90.88	&	80.02		&	0.2286 	&	0.2388 \\
    &		lstm		&	rnn-cnn	&	+ att. + weighted	&	90.88	&	83.82		&	0.2304 	&	0.2369 \\
    \bottomrule
  \end{tabular}}
     \vspace{-4mm}
\end{table*}

{\bf Corpus-based evaluation} \hspace{2mm}
We evaluated the end-to-end system by first performing a corpus-based evaluation in which the model is used to predict each system response in the held-out test set.
Three evaluation metrics were used: BLEU score (on top-1 and top-5 candidates)~\cite{papineni2002bleu}, entity matching rate and objective task success rate~\cite{SuVGKMWY15}.
We calculated the entity matching rate by determining whether the actual selected entity at the end of each dialogue matches the task that was specified to the user.
The dialogue is then marked as successful if both (1) the offered entity matches, and (2) the system answered all the associated information requests (e.g. {\it what is the address?}) from the user.
We computed the BLEU scores on the template-like output sentences before lexicalising with the entity value substitution. 

Table~\ref{tab:main} shows the result of the corpus-based evaluation averaging over 5 randomly initialised networks.
The {\it Baseline} block shows two baseline models: the first is a simple turn-level sequence to sequence model~\cite{SutskeverVL14} while the second one introduces an additional recurrence to model the dependency on the dialogue history following Serban et al~\cite{SerbanSBCP15}.
As can be seen, incorporation of the recurrence improves the BLEU score.
However, baseline task success and matching rates cannot be computed since the models do not make any provision for a database. 

The {\it Variant} block of Table~\ref{tab:main} shows two variants of the proposed end-to-end model. 
For the first one, no requestable trackers were used, only informable trackers.  Hence, the burden of modelling user requests falls on the intent network alone. 
We found that without explicitly modelling user requests, the model performs very poorly on task completion ($\sim30$\%), even though it can offer the correct entity most of the time($\sim90$\%).
More data may help here; however, we found that the incorporation of an explicit internal semantic representation in the full model (shown below) is more efficient and extremely effective.
For the second variant, the LSTM intent network is replaced by a CNN. 
This achieves a very competitive BLEU score but task success is still quite poor ($\sim58$\% success).
We think this is because the CNN encodes the intent by capturing several local features but lacks the global view of the sentence, which may easily result in an unexpected overfit.

The {\it Full model} block shows the performance of the proposed model with different decoding strategies. 
The first row shows the result of decoding using the average likelihood term (Equation~\ref{eq:ml}) while the second row uses the {\it weighted} decoding strategy (Equation~\ref{eq:weighted}). 
As can be seen, the {\it weighted} decoding strategy does not provide a significant improvement in BLEU score but it does greatly improve task success rate ($\sim3$\%).
The $R_t$ term contributes the most to this improvement because it injects additional task-specific information during decoding. 
Despite this, the most effective and elegant way to improve the performance is to use the attention-based mechanism ({\it +att.}) to dynamically aggregate the tracker beliefs (Section~\ref{ssec:gen}).
It gives a slight improvement in BLEU score ($\sim0.01$) and a big gain on task success ($\sim5$\%).
Finally, we can improve further by incorporating {\it weighted} decoding with the attention models ({\it+ att. + weighted}).

\begin{figure*}[t]
\centerline{\includegraphics[width=95mm]{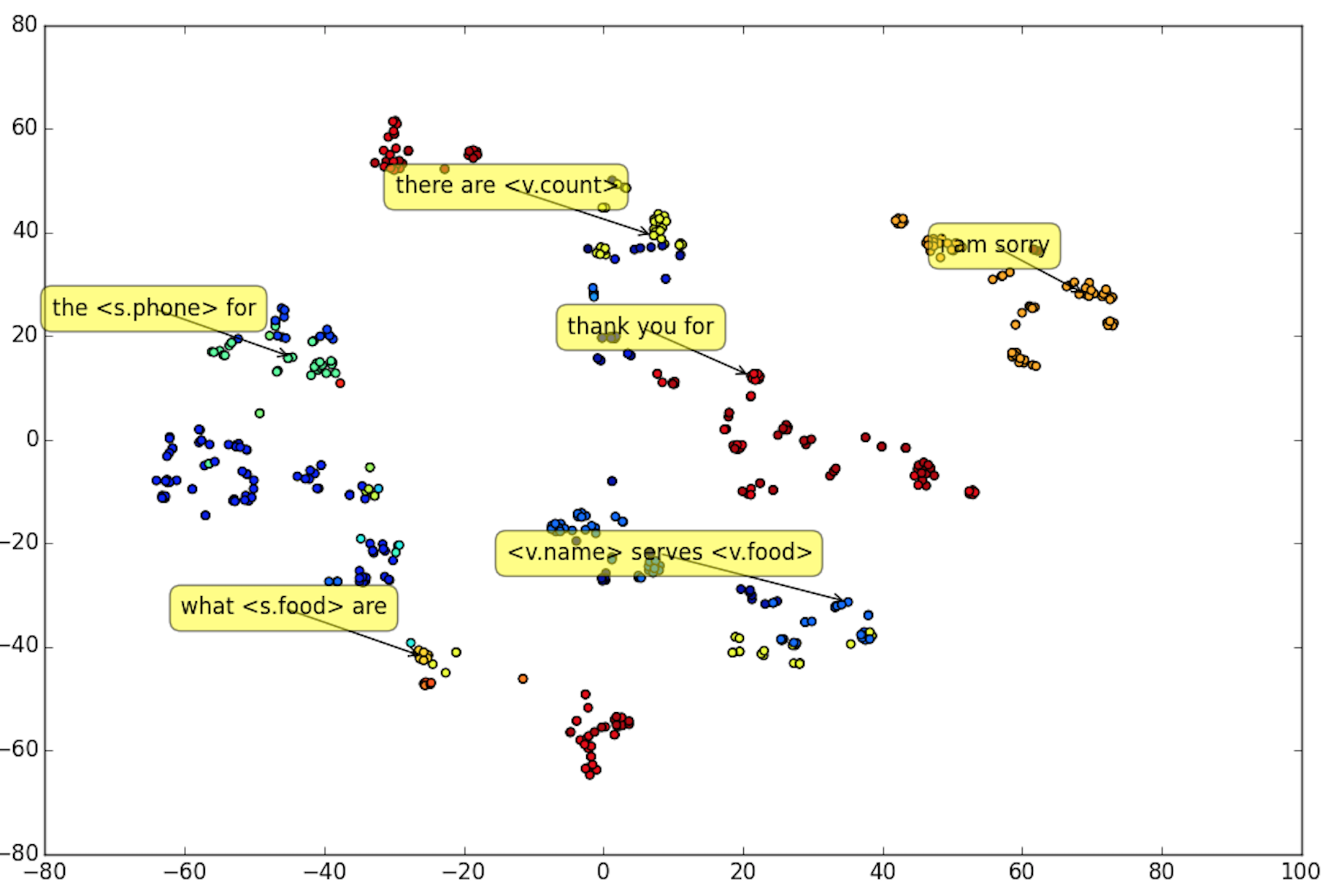}}
\caption{The action vector embedding $\mathrm{\mathbf{o}}_{t}$ generated by the NN model w/o attention. Each cluster is labelled with the first three words the embedding generated. }
\label{fig:emb}
\vspace{-3mm}
\end{figure*}

As an aside, we used t-SNE~\cite{citeulike3749741} to produce a reduced dimension view of the action embeddings $\mathrm{\mathbf{o}}_{t}$, plotted and labelled by the first three generated output words (full model w/o attention). The figure is shown as Figure~\ref{fig:emb}.
We can see clear clusters based on the system intent types, even though we did not explicitly model them using dialogue acts.

{\bf Human evaluation} \hspace{2mm}
In order to assess operational performance, we tested our model using paid subjects recruited via Amazon Mechanical Turk.
Each judge was asked to follow a given task and to rate the model's performance.
We assessed the subjective success rate, 
and the perceived comprehension ability and naturalness of response on a scale of 1 to 5.
The full model with attention and weighted decoding was used and the system was tested on a total of 245 dialogues. As can be seen in Table~\ref{tab:raw}, the average subjective success rate was 98\%, which means the system was able to complete the majority of tasks.
Moreover, the comprehension ability and naturalness scores both averaged more than 4 out of 5. (See Appendix for some sample dialogues in this trial.)

\begin{table}[t]
	\caption{Human assessment of the NN system. The rating for comprehension/naturalness are both out of 5.}
  	\label{tab:raw}
  	\centering
  	\Scale[1]{\begin{tabular}{lc}
    		\toprule
    		Metric		&	NN\\
    		\midrule
    		Success		&	98\%\\
    		\midrule
            Comprehension	&	4.11\\
    		Naturalness	&	4.05\\
    		\midrule
		  \# of dialogues:  & 245
  		\end{tabular}
	}
\end{table}

We also ran comparisons between the NN model and a handcrafted, modular baseline system ({\it HDC}) consisting of a handcrafted semantic parser, rule-based policy and belief tracker, and a template-based generator.
The result can be seen in Table~\ref{tab:comp}.
The HDC system achieved $\sim95$\% task success rate, which suggests that it is a strong baseline even though most of the components were hand-engineered.
Over the 164 dialogues tested, the NN system ({\it NN}) was considered better than the handcrafted system ({\it HDC}) on all the metrics compared.
Although both systems achieved similar success rates, the NN system ({\it NN}) was more efficient and provided a more engaging conversation (lower turn number and higher preference).
Moreover, the comprehension ability and naturalness of the NN system were also rated higher, which suggests that the learned system was perceived as being more natural than the hand-designed system.

\section{Conclusions and Future Work}\label{sec:conclusion}

\begin{table}[t]
 	\caption{A comparison of the NN system with a rule-based modular system ({\it HDC}).}
  	\label{tab:comp}
  	\centering
  	\Scale[0.9]{\begin{tabular}{llccc}
    		\toprule
    		&Metric			&	NDM		&	HDC	&	Tie\\
    		\midrule
    		&Subj. Success	&	96.95\%	&	95.12\%	&	-\\
            &Avg. \# of Turn&	3.95	&	4.54	&	-\\
    		\midrule
		\multicolumn{5}{l}{\bf Comparisons(\%)}\\
    		&Naturalness		&	46.95\tmark[{\makebox[0pt][l]{*}}] 	&	25.61	&	27.44\\
    		&Comprehension	&	45.12\tmark[{\makebox[0pt][l]{*}}]	&	21.95		&	32.93\\
    		&Preference		&	50.00\tmark[{\makebox[0pt][l]{*}}]	&	24.39	&	25.61\\
    		&Performance		&	43.90\tmark[{\makebox[0pt][l]{*}}]	&	25.61	&	30.49\\
    		\bottomrule
    		\multicolumn{5}{l}{* p \textless 0.005, ~~~\# of comparisons: 164}\\
  		\end{tabular}
	} 
	\vspace{-3mm}
\end{table}

This paper has presented a novel neural network-based framework for task-oriented dialogue systems.
The model is end-to-end trainable using two supervision signals and a modest corpus of training data.
The paper has also presented a novel crowdsourced data collection framework inspired by the Wizard-of-Oz paradigm.
We demonstrated that the pipe-lined parallel organisation of this collection framework enables good quality task-oriented
dialogue data to be collected quickly at modest cost.

The experimental assessment of the NN dialogue system showed that the learned model can interact efficiently and naturally with human subjects to complete an application-specific task.
To the best of our knowledge, this is the first end-to-end NN-based model that can conduct meaningful dialogues in a task-oriented application.

However, there is still much work left to do. 
Our current model is a text-based dialogue system, which can not directly handle noisy speech recognition inputs nor can it ask the user for confirmation when it is uncertain.
Indeed, the extent to which this type of model can be scaled to much larger and wider domains remains an open
question which we hope to pursue in our further work.

\newpage
\onecolumn

\noindent\begin{minipage}{\linewidth}

\section*{Wizard-of-Oz data collection websites} \label{app:woz}
\centerline{\includegraphics[width=150mm]{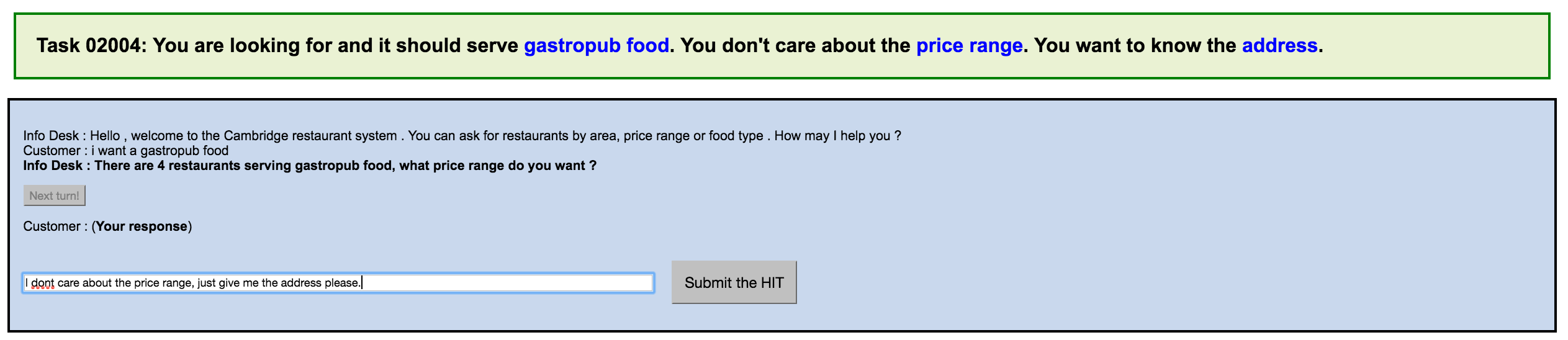}}
\captionof{figure}{The user webpage. The worker who plays a user is given a task to follow. For each mturk HIT, he/she needs to type in an appropriate sentence to carry on the dialogue by looking at both the task description and the dialogue history.}
\label{fig:wozu}

\vspace{5mm}
\centerline{\includegraphics[width=150mm]{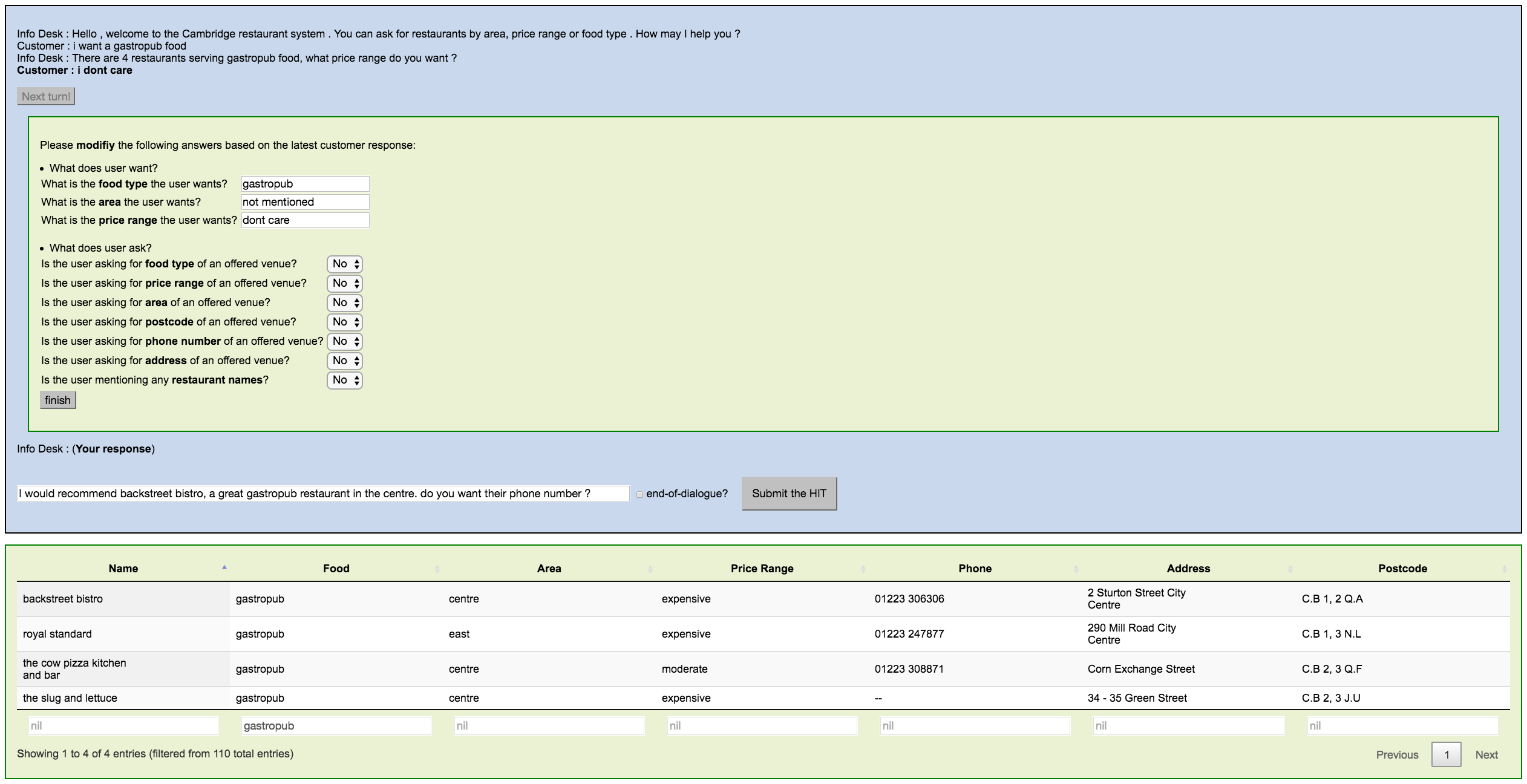}}
\captionof{figure}{The wizard page. The wizard's job is slightly more complex: the worker needs to go through the dialogue history, fill in the form (top green) by interpreting the user input at this turn, and type in an appropriate response based on the history and the DB result (bottom green). The DB search result is updated when the form is submitted. The form can be divided into informable slots (top) and requestable slots (bottom), which contains all the labels we need to train the trackers.}
\label{fig:wozw}

\vspace{5mm}
\section*{Scoring Table} \label{app:rtable}
  \captionof{table}{Additional $R_t$ term for delexicalised tokens when using weighted decoding (Equation~\ref{eq:weighted}). {\it Not observed} means the corresponding tracker has a highest probability on either {\it not mentioned} or {\it dontcare} value, while {\it observed} mean the highest probability is on one of the categorical values. A positive score encourages the generation of that token while a negative score discourages it.}
  \vspace{3mm}
  \label{tab:R}
  \centering
  \begin{tabular}{llcc}
  \toprule
  Delexicalised token		&	Examples				&	$R_t$ ({\it observed})	&	$R_t$ ({\it not observed})\\
  \midrule
  informable slot token 	& 	<s.food>, <s.area>,...	&		0.0		&		0.0		\\
  informable value token 	& 	<v.food>, <v.area>,...	&		+0.05	&		-0.5		\\
  requestable slot token 	& 	<s.phone>,<s.address>,...	&		+0.2		&		0.0		\\
  requestable value token 	& 	<v.phone>,<v.address>,...	&		+0.2		&		0.0		\\
  \bottomrule
  \end{tabular}
\end{minipage} 

\twocolumn
\section*{Acknowledgements}
Tsung-Hsien Wen and David Vandyke are supported by Toshiba Research Europe Ltd, Cambridge. The authors would like to thank Ryan Lowe and Luk\'a{\v{s}} {\v{Z}}ilka for their valuable comments.

\bibliography{eacl2017}
\bibliographystyle{eacl2017}

\appendix
\onecolumn

\noindent\begin{minipage}{\linewidth}
\section*{Appendix: Sample dialogues} \label{app:sample}
\vspace{-3mm}
  \captionof{table}{Some samples of real conversational logs between online judges and the end-to-end system.}
  \label{tab:sample}
  \centering
  \Scale[0.65]{
  \begin{tabular}{ll}
    \toprule
    Human	&: Restaurant in any area that serves Vietnamese food\\
Machine	&: thanh binh serves vietnamese food type , is there anything else i can help you with ?\\
Human	&: What is the phone number and post code\\
Machine	&: the number is 01223 362456 and the area code is c.b 3, 0 a.f . is there anything else i can do for you ?\\
Human	&: Good bye\\
Machine	&: thank you for using the cambridge restaurant system .\\    
\midrule
Human	&: i am looking for moderately priced halal food\\
Machine	&: i am sorry , there are no halal restaurants in the moderate price range . would you like something else ?\\
Human	&: how about european food\\
Machine	&: galleria serves european food type and is in the moderate price range .\\
Human	&: address and postcode\\
Machine	&: the address is 33 bridge street and the area code is c.b 2, 1 u.w . is there anything else i can help you with today ?\\
Human	&: thanks\\
Machine	&: galleria is a european restaurant in the moderate price range . would you like their phone ?\\
Human	&: thank you\\
Machine	&: thank you for using the cambridge restaurant system . good bye\\
\midrule
Human	&: Hi, I'm hungry for some Indonesian food\\
Machine	&: i am sorry , there are no indonesian restaurants . would you like a different food ?\\
Human	&: Oh no. How about Indian?\\
Machine	&: there are 9 restaurants serving indian types of food . do you have a preference on side of town ?\\
Human	&: East side please\\
Machine	&: sitar tandoori is a indian restaurant in the east side of town . would you like their information ?\\
Human	&: Yes, the address and phone number please\\
Machine	&: sitar tandoori is located at 43 high street cherry hinton cherry hinton and their phone number is 01223 249955 .\\
Human	&: Okay, thank you\\
Machine	&: thank you , goodbye .\\
\midrule
Human	&: hello\\
Machine	&: what cuisine would you like ?\\
Human	&: afghan\\
Machine	&: i am sorry , there are no afghan restaurants . would you like something else ?\\
Human	&: korean\\
Machine	&: little seoul serves korean food , it is located at 108 regent street city centre .\\
Human	&: is it expensive?\\
Machine	&: little seoul serves korean food and is in the expensive price range .\\
Human	&: what area is that\\
Machine	&: little seoul is located in the centre area of town . would you like their number ?\\
Human	&: yes\\
Machine	&: the telephone number is 01223 308681 . is there anything else i can help you with today ?\\
Human	&: no goodbye\\
Machine	&: thank you for using the cambridge restaurant system , goodbye .\\
\midrule
Human	&: I want to find a restaurant in the east part of town that serves creative food\\
Machine	&: i am sorry , there are no creative restaurants in the east side of town . would you like a different restaurant ?\\
Human	&: how about indian food?\\
Machine	&: curry prince is a indian restaurant in the east of town . would you like their phone number ?\\
Human	&: yes, may i have the address and phone number?\\
Machine	&: curry prince is located at 451 newmarket road fen ditton and their phone is 01223 566388 . is there anything else i can help you with today ?\\
Human	&: no, thank you goodbye\\
Machine	&: thank you for using the cambridge restaurant system . good bye\\
\midrule
Machine	&: what food are you interested ?\\
Human	&: english\\
Machine	&: i am sorry , there are no english restaurants . would you like to try a different food type ?\\
Human	&: chinese\\
Machine	&: there are 8 restaurants serving chinese type of cuisine . do you have a preference of area ?\\
Human	&: south part\\
Machine	&: the good luck chinese food takeaway is a chinese restaurant in the south of town . would you like the phone number ?\\
Human	&: no. just address and price range\\
Machine	&: the good luck chinese food takeaway is located at 82 cherry hinton road cherry hinton . is there anything else i can help you with ?\\
Human	&: what is their price range?\\
Machine	&: the good luck chinese food takeaway is expensively priced chinese restaurant in the south area . would you like more information ?\\
Human	&: no goodbye\\
Machine	&: thank you for using the cambridge restaurant system . good bye .\\
\midrule
Human	&: Is there a gastropub anywhere around here?\\
Machine	&: there are 3 gastropub restaurants . would you like more information ?\\
Human	&: Is there one on the west side?\\
Machine	&: i am sorry , there are no gastropub restaurants in the west side of town . would you like something else ?\\
Human	&: How about east?\\
Machine	&: royal standard is a gastropub restaurant in the east of town .\\
Human	&: What's the address?\\
Machine	&: the address is 290 mill road city centre .\\
Human	&: Thanks, bye\\
Machine	&: thank you for using the cambridge restaurant system . good bye\\
\bottomrule
  \end{tabular}}
\end{minipage} 

\end{document}